\documentclass[10pt,twocolumn,letterpaper]{article}

\usepackage{cvpr}              % To produce the CAMERA-READY version
\usepackage[utf8]{inputenc}
\usepackage{amsmath}
\usepackage{algorithm}
\usepackage{algpseudocode}
\usepackage{amssymb}   
\usepackage{multirow}
\usepackage{comment}

\newcommand{\worldwideweb}{\raisebox{-1.5pt}{\includegraphics[height=1.05em]{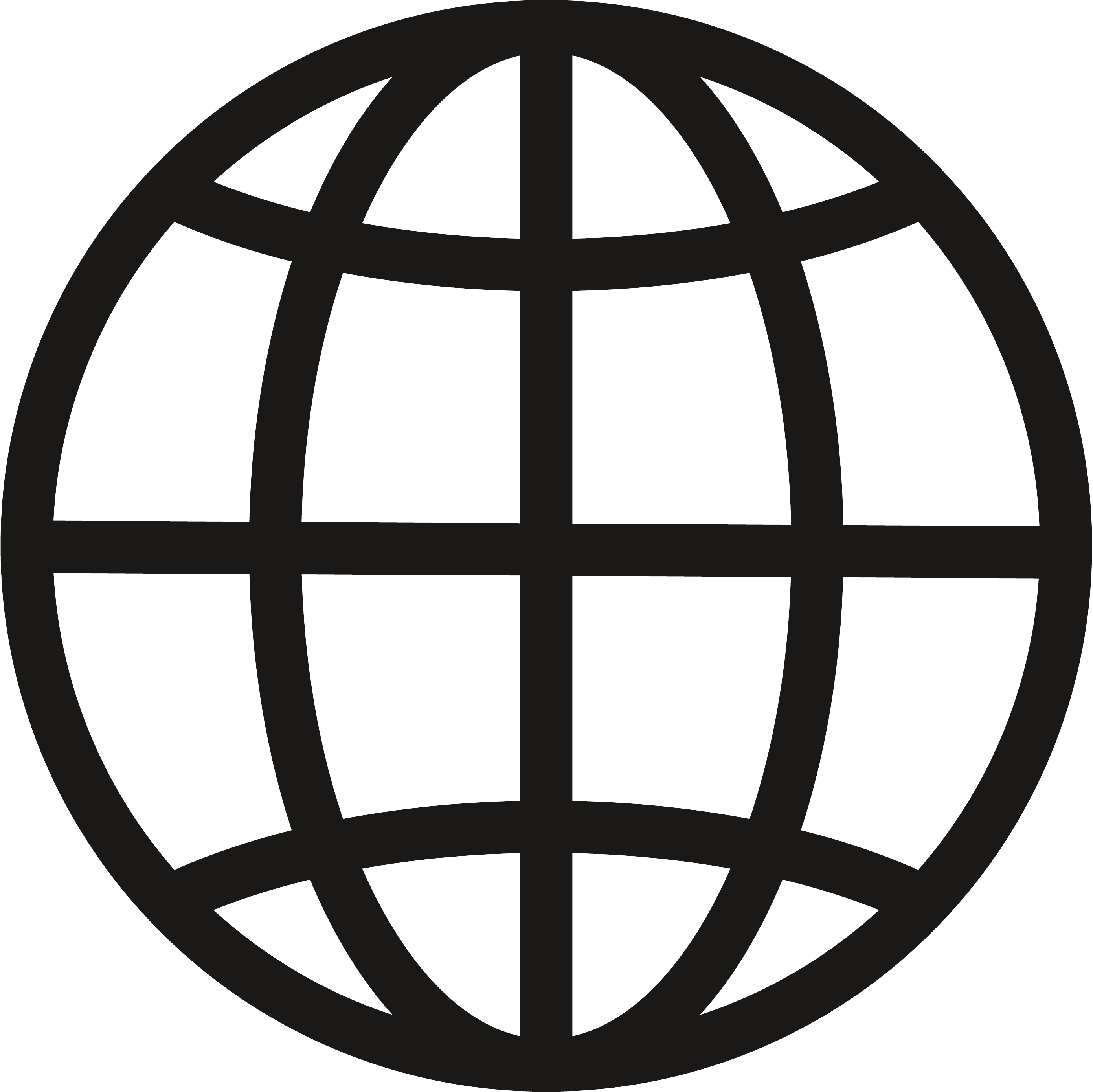}}\xspace}
\newcommand{\github}{\raisebox{-1.5pt}{\includegraphics[height=1.05em]{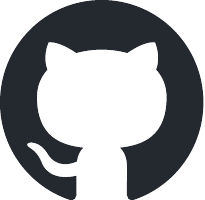}}\xspace}

\definecolor{cvprblue}{rgb}{0.21,0.49,0.74}
\usepackage[pagebackref,breaklinks,colorlinks,allcolors=cvprblue]{hyperref}

\def\confName{CVPR}
\def\confYear{2026}

\title{DecoVLN: Decoupling Observation, Reasoning, and Correction \\ for Vision-and-Language Navigation}

\author{
Zihao Xin$^1$\footnotemark[1], \ 
Wentong Li$^{1*,\dagger}$, \ 
Yixuan Jiang$^1$,  \   
Bin Wang$^2$, \   
Runmin Cong$^2$, \    
Jie Qin$^1$\footnotemark[3], \ 
Shengjun Huang$^1$\footnotemark[3] \\[0.2cm]
$^1$Nanjing University of Aeronautics and Astronautics \ \  \ \  
$^2$Shandong University 
\\
[0.2cm]
{\worldwideweb \href{https://allenxinn.github.io/DecoVLN/\#}{{\text{Project Page}}}} \quad \quad {\github \href{https://github.com/Allenxinn/DecoVLN}{{\text{Code}}}}
}

\begin{document}
\maketitle

\renewcommand{\thefootnote}{\fnsymbol{footnote}}
\footnotetext[1]{Equal contribution}
\footnotetext[2]{Project lead}
\footnotetext[3]{Corresponding author} 
\begin{abstract}
 Vision-and-Language Navigation (VLN)  requires agents to follow long-horizon instructions and navigate complex 3D environments. However, existing approaches face two major challenges: constructing an effective long-term memory bank and overcoming the compounding errors problem. 
 To address these issues, we propose DecoVLN, an effective framework designed for robust streaming perception and closed-loop control in long-horizon navigation. 
 First, we formulate long-term memory construction as an optimization problem and introduce adaptive refinement mechanism that selects frames from a historical candidate pool by iteratively optimizing a unified scoring function.  
 This function jointly balances three key criteria: semantic relevance to the instruction, visual diversity from the selected memory, and temporal coverage of the historical trajectory. Second, to alleviate compounding errors, we introduce a state-action pair-level corrective  finetuning strategy. 
 By leveraging geodesic distance between states to precisely quantify deviation from the expert trajectory, the agent  collects high-quality state-action pairs in the trusted region while filtering out the polluted data with low relevance. This improves both the efficiency and stability of error correction. Extensive experiments demonstrate the effectiveness of  DecoVLN, and we have deployed it in real-world environments. 
\end{abstract}    
\section{Introduction}
\label{sec:intro}

Vision-and-Language Navigation (VLN) has emerged as a fundamental task in Embodied AI, requiring an autonomous agent 
to interpret natural language instructions and navigate through complex, previously unseen environments based solely on egocentric visual observations~\cite{zhang2024navid, zhang2024uninavid, cheng2024navila, streamvln}.
Achieving success requires not only high-level language understanding and precise visual grounding, but also the ability to perform robust spatial and temporal reasoning over long-horizon decision-making.

Existing approaches can be broadly categorized into two paradigms:  
(\textit{i}) \textbf{Stop-and-Think}, where the agent executes an action and then pauses to perceive and reason before proceeding~\cite{zhang2024navid, cheng2024navila}. This low-frequency, discontinuous perception mode often leads to \textit{perceptual blindness}, missing critical visual cues during movement. 
(\textit{ii}) \textbf{Full-History Streaming}, which continuously appends all observed frames to the history sequence, with contextual retrieved via fixed sampling or heuristic strategies~\cite{vln_r1, streamvln}. However, this  often dilutes the density of contextual observations, thereby impairing long-horizon reasoning.
Moreover, as a sequential decision-making task, VLN is highly susceptible to \textit{compounding errors}~\cite{zhang2024uninavid, vln_r1, cheng2024navila}. 
Even minor early-stage action errors can accumulate over time,  leading the agent to deviate significantly from the intended path.  To mitigate this, most methods~\cite{streamvln, cheng2024navila, zhang2024uninavid} focus on multimodal trajectory augmentation to improve open-loop action prediction. However, they often lack effective closed-loop reflection and online  correction, leaving agents poorly equipped to recover once off-course.

To address these challenges, we propose \textbf{DecoVLN}, an effective VLN framework for robust streaming perception and closed-loop control.  DecoVLN tackles long-horizon navigation by explicitly decoupling observation, reasoning and correction. 
We begin by formulating VLN as a Partially Observable Markov Decision Process (POMDP)~\cite{pomdp}, where the agent lacks access to a global map and must infers a belief state based on observation history $H_t$ and instruction $I$ to learn a policy $\pi(H_t, I)$ that maximizes expected cumulative reward. This formulation poses \textbf{three core challenges}: managing the dynamic memory, performing efficient action reasoning, and enabling timely policy correction.
To this end, we introduce two key designs: an adaptive memory refinement mechanism and a state–action pair-based corrective fine-tuning strategy.

Specifically, we first disentangle the agent’s observation stream from its reasoning stream, allowing continuous perception in parallel with action execution. The proposed adaptive memory refinement mechanism dynamically assesses each incoming observation based on its semantic relevance, visual diversity, and temporal coverage with respect to the current navigation sub-goal. 
By filtering out redundant or low-value frames, the agent maintains a compact, high-information-density memory representation, thereby mitigating context interference and enhancing long-horizon reasoning.
Beyond visual perception, we further introduce a corrective fine-tuning strategy that actively collects high-quality state–action pairs for policy refinement. This strategy empowers the agent to reflect on and correct its deviations, significantly reducing compounding errors and improving both policy robustness and data efficiency.

Extensive experiments on classic VLN benchmarks demonstrate that DecoVLN consistently outperforms prior state-of-the-art methods, validating its effectiveness in both perception and control. Furthermore, we deploy DecoVLN in real-world scenarios, where it exhibits strong robustness and generalization when following complex instructions across diverse environments. 
In summary, our main contributions are threefold:
\begin{itemize}
    \item We propose DecoVLN, an effective VLN framework  that decouples observation, reasoning and correction, enabling robust long-horizon navigation. Our approach achieves superior results on classic benchmarks.
    \item We introduce an  adaptive memory refinement  mechanism, which effectively selects frames from a historical candidate pool and substantially improving the effective information density of navigation context.
    \item We design a  state–action pair-based corrective fine-tuning strategy, which enhances the model’s multi-modal reasoning robustness and proactive error-correction capability in complex environments.
\end{itemize}

\section{Related Work}
\label{sec:related}

\subsection{Vision-Language Model}
In recent years, significant breakthroughs in text understanding and generation by Large Language Models (LLMs) have prompted researchers to explore their extension into the visual domain, giving rise to Vision–Language Large Models (VLMs)~\cite{blip_2,llava,shi2024eagle,li2025eagle,yuan2024osprey,li2025tokenpacker,yuan2025pixelrefer,wang2025videoitg,yuan2025videorefer}. Early work, such as BLIP-2~\cite{blip_2}, utilize a lightweight Query Transformer to align pre-trained visual encoders with frozen LLMs, enabling efficient cross-modal representation learning with minimal supervision. The LLaVA  series~\cite{llava, improvedllava, liu2024llavanext} is the first to systematically propose visual instruction tuning paradigm, which unifies visual perception capabilities and natural language reasoning abilities. Building upon this, Video-LLaVA~\cite{video_llava} extends the framework to the video domain by incorporating frame-wise visual encoding and temporal attention mechanisms. 
The Qwen-VL series~\cite{qwen2_vl, qwen3, qwen_image} propose dynamic frame sampling and visual token compression, offering a generalizable framework for both video and multi-image understanding.
Despite these advancements, existing studies~\cite{gholami2025spatial, qi2025gpt4scene, yu2025inst3d, yuan2025eoc} have revealed that current VLMs exhibit limited 3D spatial understanding.
This is primarily due to the model's inability in establishing correspondence between local egocentric observations with a coherent global spatial structure~\cite{wang2025scalable},  a capability that is crucial for navigation tasks.

\subsection{Vision-and-Language Navigation}
The VLN task requires an autonomous agent to navigate   3D environments based on natural language instructions~\cite{streamvln,zhang2024navid,yu2025correctnav,wei2025ground,xue2025omninav,yuan2025multimodal}. 
Early methods~\cite{seq2seq21, seq2seq22, seq2seq23} primarily employ a Seq2Seq~\cite{seq2seq} architecture, mapping language instructions to action sequences via imitation learning. 
With the emergence of LLMs exhibiting strong general reasoning capabilities, methods such as NavGPT~\cite{navgpt} and InstructNav~\cite{instructnav} leverage LLMs for path planning in a zero-shot manner. 
However, these approaches typically require additional  perception modules to convert raw visual inputs into textual descriptions. In contrast, NaVid~\cite{zhang2024navid} is the first to directly fine-tune a VLM for action prediction, enabling end-to-end navigation. 
Recent efforts have focused on two major challenges. The first is the lack of 3D spatial understanding, which is often mitigated through large-scale multi-modal dataset augmentation to implicitly enhance spatial reasoning. 
The second is the context length bottleneck in long-horizon navigation, where uniform sampling strategies are commonly employed to capture global observations~\cite{zhang2024navid, vln_r1, streamvln, cheng2024navila}. 
However, such approaches  discard critical navigation node cues and disrupt temporal coherence.
StreamVLN~\cite{streamvln}  introduces a Slow-Fast memory mechanism that combines sliding temporal windows and voxel-based memory compression to mitigate this issue, though it depends on depth sensors for voxel construction.

\section{Method}
\label{sec:method}
\begin{figure*}[]
  \centering
   \includegraphics[width=1.0\linewidth]{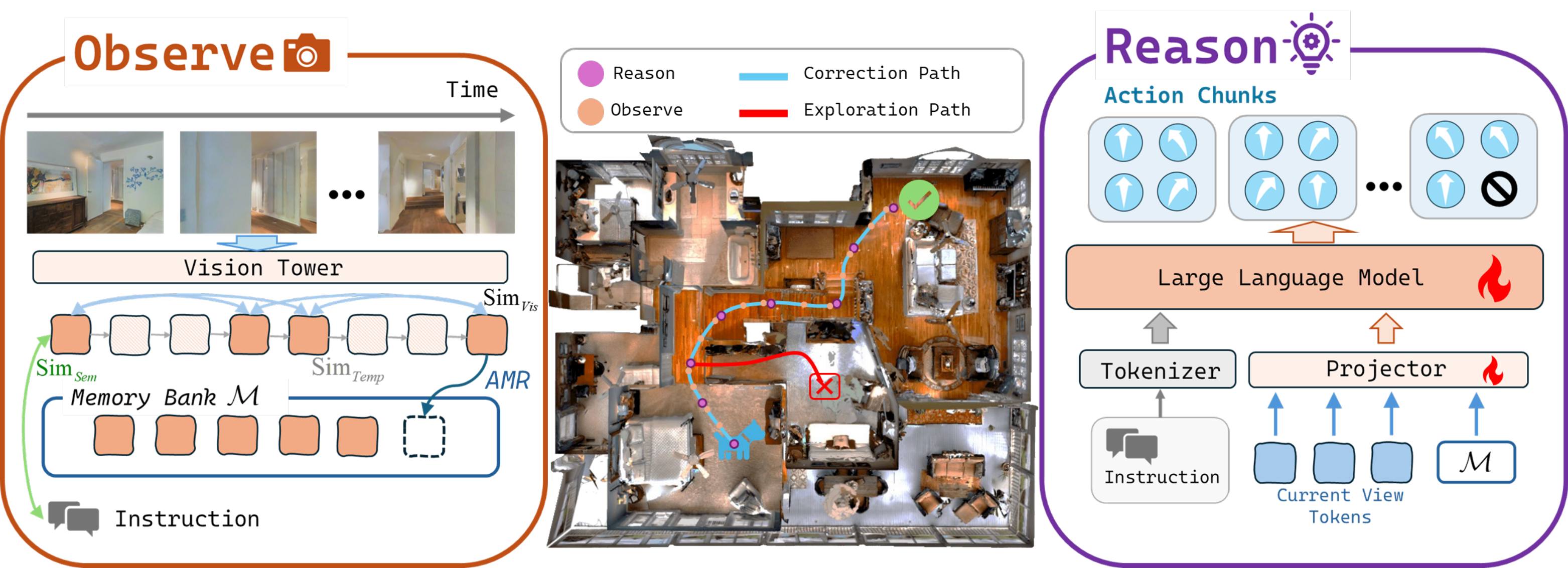}
   \caption{The framework of DecoVLN. DecoVLN decouples the agent's observation and reasoning processes. The agent can perceive the environment continuously while in motion and, based on the Adaptive Memory Refinement (AMR) mechanism, it filters and stores high-information-density state representations into a memory bank. During the generation phase, the LLM outputs the action chunk which is comprising multiple consecutive actions-based on the input instruction, the current frame, and the memory bank. Subsequently, we construct an error-correction strategy based on state-action pairs. The model autonomously explores according to the instruction and collects State-Action Pairs within a trusted region for error-correction fine-tuning. This process not only enhances data utilization efficiency but also equips the model with introspective and self-correction capabilities.}
   \label{fig:frameword}
   \vspace{-0.2cm}
\end{figure*}

In this section, we present DecoVLN, a framework that decouples observation, reasoning and correction for long-horizon navigation.  We first formulate  the VLN problem as a long-horizon optimization task. To address its core challenges: efficient memory construction and robust closed-loop control, we introduce two key components: an adaptive memory refinement mechanism and a state–action pair-based corrective fine-tuning strategy.
The overall framework of our method is illustrated in ~\cref{fig:frameword}.

% \subsection{Observation and Reasoning}
\subsection{Problem Formulation}
VLN can be formulated as a Partially Observable Markov Decision Process (POMDP), which evaluates an embodied agent’s ability to ground a language instruction $I$ within its egocentric visual observations to achieve long-horizon navigation goals. 
A POMDP is defined as a tuple $M = (S, A, T, R, \Omega, O)$, where $S$ represents the set of all possible true states $s$, which are not directly observable. $A$ denotes the discrete action space, typically including $\{\texttt{MOVE\_FORWARD}, \texttt{TURN\_LEFT}, \texttt{TURN\_RIGHT}, \texttt{STOP}\}$ in continuous environments. $T$ defines the state transition function, $R$ is the reward function, $\Omega$ represents the observation space of egocentric RGB images $f_t$, where the subscript $t$ denotes the time step, and $O$ is the observation function. 
The reward $R$ is general sparse, granting a large positive reward when the agent executes the `\texttt{STOP}' action within the target area, and small negative penalties at each step to encourage efficiency. Since the true state $s$ is unobservable, the agent must learn a policy $\pi$ conditioned on its observation history $H_t = (O_1, a_1, O_2, a_2, ..., O_t)$ and the instruction $I$.  The policy $\pi(H_t, I)$ aims to maximize the expected cumulative reward $E[\sum_{t=0}^{T} \gamma^t R_t]$. Here, $\gamma \in [0, 1]$ serves as the discount factor for time preference. This implies that in the VLN task, higher weight is assigned to immediate rewards compared to distant future rewards, which helps to balance the value of short-term rewards against the ultimate goal.

\subsection{Observation and Reasoning}
% We propose a framework which decouples observation and reasoning, designed 
To mitigate issues such as perceptual blindness, we first decouple the processes of observation and reasoning. During navigation, the agent continuously perceives the environment and evaluates each new observation frame online through the \textbf{Adaptive Memory Refinement} module.
This module actively filters out redundant or weakly relevant frames based on feature novelty and semantic relevance. 
The resulting memory stream exhibits a higher signal-to-noise ratio, enabling a fixed-length memory window to retain semantically richer and temporally broader context. This prevents information overload and enhances the agent’s capacity for long-horizon reasoning.

% This process enhances the signal-to-noise ratio of contextual features, prevents the agent from being overwhelmed by redundant information, and allows a fixed-length memory window to contain semantically richer, temporally extended history. 

% \subsection{Adaptive Memory Refinement}
Unlike existing methods that rely on fixed sampling or spatial redundancy pruning, we argue that an ideal memory buffer $\mathcal{M}$ should simultaneously satisfy three key properties: semantic relevance, visual diversity, and temporal coverage. We formulate the construction of long-term memory as an optimization problem. During navigation, the agent continuously collects egocentric visual frames. At each timestep $t$, it iteratively selects $K$ frames from a candidate pool $\mathcal{C}$ to form a refined memory $\mathcal{M}$, maximizing an overall scoring objective:
\begin{equation}
\begin{aligned}
f^* &= \arg\max_{f \in \mathcal{C} \setminus \mathcal{M}} 
[ 
\lambda_R \cdot \mathrm{Sim}_{\mathrm{Sem}}(f, I)
- (1 - \lambda_R) \cdot \\ 
&\left( 
w_V \cdot \mathrm{Sim}_{\mathrm{Vis}}(f, \mathcal{M}) 
+ w_T \cdot \mathrm{Sim}_{\mathrm{Temp}}(f, \mathcal{M})
\right)],
% \right]
\end{aligned}
\end{equation}
where $\mathcal{M}$ is the current set of selected memory frames, $f^*$ denotes the optimal frame to be added. The scalar $\lambda_R \in [0, 1]$ controls the trade-off between instruction relevance and redundancy, while $w_V$ and $w_T$ are weighting coefficients that balance visual and temporal redundancy, respectively, subject to the constraint $w_V + w_T = 1$.

\noindent\textbf{Semantic Relevance.} We employ an efficient \textit{Instruction–Image Feature Alignment} mechanism. The language instruction $I$ is encoded into a global embedding vector $\mathbf{e}_I \in \mathbb{R}^d$ using the text encoder of a VLM. For each candidate frame $f$ in $\mathcal{C}$, a visual embedding $\mathbf{e}_f \in \mathbb{R}^d$ is obtained via a visual encoder followed by multimodal projection layer. The semantic relevance score $\text{Sim}_{\text{Sem}}(f, I)$ is defined as the cosine similarity between the two embeddings:
\begin{equation}
\text{Sim}_{\text{Sem}}(f, I) = \text{sim}(\mathbf{e}_f, \mathbf{e}_I)
= \frac{\mathbf{e}_f \cdot \mathbf{e}_I}{\|\mathbf{e}_f\| \, \|\mathbf{e}_I\|}.
\end{equation}While focusing solely on relevance may cause the memory to concentrate on a few key scenes, we additionally incorporate visual diversity and temporal coverage to ensure a more comprehensive and balanced historical representation.

\noindent\textbf{Visual Diversity.} To promote diversity, we penalize candidate frames that are visually too similar to those already in memory buffer $\mathcal{M}$. Leveraging the feature embeddings extracted by the visual encoder, we define the visual similarity penalty as the maximum cosine similarity between a candidate frame $f$ and any memory frame $m \in \mathcal{M}$:
\begin{small}
\begin{equation}
\text{Sim}_{\text{Vis}}(f, \mathcal{M}) =
\max_{m \in \mathcal{M}}
\left(
\frac{\text{embed}(f) \cdot \text{embed}(m)}
{\|\text{embed}(f)\| \, \|\text{embed}(m)\|}
\right).
\end{equation}
\end{small}

\noindent\textbf{Temporal Coverage.} To ensure temporal coverage, we penalize candidate frames that are temporally close to those already stored in the memory buffer $\mathcal{M}$.
% are temporally too close to existing memory frames. 
The temporal similarity term is defined as:
\begin{equation}
\text{Sim}_{\text{Temp}}(f, \mathcal{M}) =
\frac{1}{\min_{m \in \mathcal{M}} |t_f - t_m| + \epsilon},
\end{equation}
where $t_f$ and $t_m$ denote the timestamps of the candidate frame $f$ and memory frame $m$, respectively,
% where $t_f$ and $t_m$ are timestamps of the candidate and memory frames, respectively, 
and $\epsilon$ is a small constant to avoid division by zero. This formulation encourages the selection of frames that are temporally distant from those already in memory, thereby ensuring broader temporal coverage along the navigation trajectory.
% This term encourages the selection of frames temporally distant from those already stored, promoting broader temporal coverage across the trajectory.

In our DecoVLN framework, the agent maintains continuous observation of the environment while effectively addressing the data deluge caused by high-frequency perception, significantly increasing the contextual information density.  As a result, the model achieves stronger long-horizon reasoning and memory capabilities.

\begin{algorithm}[t]
\caption{Correction Data Collection}
\label{alg:correction}
\begin{algorithmic}[1]

\Require Initial policy $\pi_\theta$, expert policy $\pi^*$,
         dataset $\mathcal{D}=\{(P_{\mathrm{exp}}^i, I^i)\}_{i=1}^M$,
         deviation threshold $\tau$
\Ensure Correction dataset $\mathcal{D}_c$

\State $\mathcal{D}_c \gets \varnothing$
      \Comment{\textcolor{gray}{\small Initialize correction dataset}}

\ForAll{episode $(P_{\mathrm{exp}}, I) \in \mathcal{D}$}
    \State $P_{\mathrm{exp}} \gets \{s_0^*, s_1^*, \dots, s_N^*\}$
           \Comment{\textcolor{gray}{\small Expert reference trajectory}}
    \State $H \gets \varnothing$
           \Comment{\textcolor{gray}{\small Initialize memory state}}

    \For{$t = 0$ \textbf{to} $T_{\max}$}
        \State \textcolor{blue}{\textbf{$DM \gets \min_{s^* \in P_{\mathrm{exp}}} d_g(s_t, s^*)$}}
        \Comment{\textcolor{gray}{\small Compute deviation from expert trajectory}}
        \If{$0 < DM(s_t) \le \tau$}
            \State \textcolor{blue}{\textbf{$a_t^{\mathrm{exp}} \gets \pi^*(s_t, P_{\mathrm{exp}})$}}
            \State \textbf{\textcolor{blue}{$\mathcal{D}_c \gets \mathcal{D}_c \cup \{(s_t, a_t^{\mathrm{exp}}, f_t)\}$}}
            \Comment{\textcolor{gray}{\small Store expert correction}}
        \ElsIf{$DM(s_t) > \tau$}
            \State \textbf{break}
            \Comment{\textcolor{gray}{\small Abort episode due to large deviation}}
        \EndIf
        \State \textcolor{blue}{$a_t \sim \pi_\theta(f_t, I, H)$}
              \Comment{\textcolor{gray}{\small Roll out current policy}}
        \State $s_{t+1} \gets \mathrm{Env.Step}(a_t)$
               \Comment{\textcolor{gray}{\small Environment state update}}
        \State $f_{t+1} \gets \mathrm{Env.Observe}(s_{t+1})$
               \Comment{\textcolor{gray}{\small Acquire next observation}}
        \State \textcolor{blue}{\textbf{$H \gets \mathrm{AMR}(H, f_{t+1})$}}
               \Comment{\textcolor{gray}{\small Adaptive memory refinement}}
        \If{target reached \textbf{or} episode failure}
            \State \textbf{break}
            \Comment{\textcolor{gray}{\small Terminate episode}}
        \EndIf

    \EndFor
\EndFor

\State \Return $\mathcal{D}_c$
 % \vspace{-0.2cm}
\end{algorithmic}
% \vspace{-0.1cm}
\end{algorithm}
% \vspace{-0.3cm}

\subsection{Corrective Fine-tuning}
In conventional VLN models, imitation learning (IL) is commonly used to improve robustness in sequential decision-making~\cite{streamvln}. During rollout, the model collects deviated states and queries an expert policy to obtain corrective actions, thereby iteratively augmenting the dataset. However, since navigation follows a Markov decision process, the agent is highly susceptible to compounding errors. Naively collected correction samples may diverge significantly from the intended instruction, 
% be poorly aligned with the original instruction,
% leading to distributional mismatch and hindering convergence.
contaminating the data distribution and impeding convergence.

To overcome this, we propose a  corrective fine-tuning strategy based on state–action pair. 
The key idea is to perform correction at the \textit{step level} rather than the \textit{episode level}, and quantify state deviations via the geometric constraints imposed by the environment, thereby maximizing both data utilization and semantic relevance.  Let the expert trajectory be  $P_{exp} = \{s_0^*, s_1^*, ..., s_N^*\}$. For each agent state $s_t$ during rollout, its deviation metric is defined as:
\vspace{-0.1cm}
\begin{equation}
DM(s_t) = \min_{s^* \in P_{exp}} d_g(s_t, s^*),
\end{equation}
where $d_g(\cdot, \cdot)$ denotes the geodesic distance between two states. The fine-tuned policy $\pi_\theta$ executes from the initial state $s_0$ and computes $DM(s_t)$ at each step. A deviation threshold $\tau$ is introduced to determine whether the current state lies within a trustworthy region. If the deviation exists but remains within this region, the expert policy $\pi^*$ is queried to obtain the corrective action $a_t = \pi^*(s_t)$. The resulting state–action pair $(s_t^{stu}, a_t^{exp}, \mathbf{f})$, where $\mathbf{f}$ is the corresponding egocentric observation, is then stored in the corrective dataset $D_c$, as detailed in the \cref{alg:correction}.

\section{Experiment}
\label{sec:exp}

\subsection{Datasets and Evaluation Metrics}
The navigation-action training dataset is collected from the Matterport3D~\cite{matterport3d} environment. The instruction and path dataset primarily consists of three subsets: R2R-CE~\cite{r2r}, R2R-Envdrop~\cite{tan2019learning}, and RxR-CE~\cite{rxr}. Each oracle trajectory contains a navigation instruction, step-wise ego-centric images, and the corresponding navigation actions. In total, we collected about 360K samples for training and employed random action execution and image augmentation strategies to enhance the model's generalization performance.

During the corrective fine-tuning stage, we introduce two additional datasets to address the problems of \emph{compounding errors} in navigation and \emph{catastrophic forgetting} in the VLM. We employ Habitat’s shortest-path follower as the expert policy $\pi^*$ to obtain corrective actions for the current states, resulting in approximately 180K corrective samples. To mitigate catastrophic forgetting and preserve the VLM’s multimodal reasoning ability acquired during pretraining, we further integrate the LLaVA-Video-178K dataset~\cite{llava_video_178k} during corrective fine-tuning. This dataset contains diverse video–language question–answer pairs, ensuring that the model improves its navigation-specific expertise without sacrificing its foundational visual–language understanding. We adopt five widely used metrics to comprehensively evaluate the agent’s navigation performance: Success Rate (SR), Success weighted by Path Length (SPL), Navigation Error (NE), Oracle Success Rate (OS), and Normalized Dynamic Time Warping (nDTW).

\subsection{Implementation Details}

For a fair comparison, our model follows previous VLN setups and is built upon the LLaVA-Video-7B\cite{llava_video} model, which uses SigLIP~\cite{siglip} as the vision tower and Qwen2-7B\cite{qwen2} as the language model. We adopt the AdamW optimizer~\cite{adamw}, with a peak learning rate of $2 \times 10^{-5}$ for the language model and $5 \times 10^{-6}$ for the vision encoder. The batch size is set to 128, and the memory bank size $K=8$, while the trust-region threshold is $\tau=3$. The model is trained on eight Nvidia A800 GPUs, with a total of approximately 600 GPU-hours. During inference, the model takes the instruction, memory bank, and the most recent four visual frames as input, and predicts an autoregressive action chunk consisting of four consecutive actions. Additional implementation details are provided in the Appendix.

\subsection{Navigation Experiments}

\begin{table*}[]
\begin{tabular}{ccccccccccccc}
\hline
\multirow{2}{*}{Method}& \multicolumn{4}{c}{Observation} & \multicolumn{4}{c}{R2R}   & \multicolumn{4}{c}{RXR}  \\ \cline{2-13}
                       & RGB   & Pano.  & Depth  & Odo.  & NE ↓ & OS↑  & SR↑  & SPL↑ & NE ↓  & SR↑  & SPL↑ & nDTW↑ \\ \hline
HPN+DN~\cite{krantz2021waypoint}                 &   & \checkmark  & \checkmark  & \checkmark & 6.31 & 40.0 & 36.0 & 34.0 & - & -& -& -     \\
VLN BERT~\cite{hong2022bridging}               &   & \checkmark  & \checkmark  & \checkmark & 5.74 & 53.0 & 44.0 & 39.0 & 8.98  & 27.0 & 22.6 & 46.7  \\
CMA~\cite{hong2022bridging}                    &   & \checkmark  & \checkmark  & \checkmark & 6.20 & 52.0 & 41.0 & 36.0 & 8.76  & 26.5 & 22.1 & 47.0  \\
Reborn ~\cite{an20221st}                &   & \checkmark  & \checkmark  & \checkmark & 5.40 & 57.0 & 50.0 & 46.0 & 5.98  & 48.6 & 42.0 & 63.3  \\
Ego$^{2}$-Map~\cite{hong2023learning} &   & \checkmark  & \checkmark  & \checkmark & 5.54 & 56.0 & 47.0 & 41.0 & - & -& -& -     \\
DreamWalker~\cite{wang2023dreamwalker}            &   & \checkmark  & \checkmark  & \checkmark & 5.53 & 59.0 & 49.0 & 44.0 & - & -& -& -     \\
HAMT+ScaleVLN~\cite{scalevln}          &   & \checkmark  & \checkmark  & \checkmark & 4.80 & -& 55.0 & 51.0 & - & -& -& -     \\
ETPNav~\cite{an2023etpnav}                 &   & \checkmark  & \checkmark  & \checkmark & 4.71 & 65.0 & 57.0 & 49.0 & 5.64  & 54.7 & 44.8 & 61.9  \\ \hline
Seq2Seq~\cite{krantz2020beyond}                & \checkmark &    & \checkmark  &   & 7.77 & 37.0 & 25.0 & 22.0 & 12.10 & 13.9 & 11.9 & 30.8  \\
RGB-CMA~\cite{krantz2020beyond}                 & \checkmark &    &    &   & 9.55 & 10.0 & 5.0  & 4.0  & - & -& -& -     \\
AG-CMTP~\cite{chen2021topological}                &   & \checkmark  & \checkmark  & \checkmark & 7.90 & 39.0 & 23.0 & 19.0 & - & -& -& -     \\
R2R-CMTP~\cite{chen2021topological}               &   & \checkmark  & \checkmark  & \checkmark & 7.90 & 38.0 & 26.0 & 22.0 & - & -& -& -     \\
LAW~\cite{raychaudhuri2021law}                    & \checkmark &    & \checkmark  & \checkmark & 6.83 & 44.0 & 35.0 & 31.0 & 10.90 & 8.0  & 8.0  & 38.0  \\
CM2~\cite{georgakis2022cross}                    & \checkmark &    & \checkmark  & \checkmark & 7.02 & 41.0 & 34.0 & 27.0 & - & -& -& -     \\
WS-MGMap~\cite{chen2022weakly}               & \checkmark &    & \checkmark  & \checkmark & 6.28 & 47.0 & 38.0 & 34.0 & - & -& -& -     \\
ETPNav+FF~\cite{wang2024sim}              & \checkmark &    & \checkmark  & \checkmark & 5.95 & 55.8 & 44.9 & 30.4 & 8.79  & 25.5 & 18.1 & -     \\
AO-Planner~\cite{chen2024affordances}             &   & \checkmark  & \checkmark  &   & 5.55 & 59.0 & 47.0 & 33.0 & 7.06  & 43.3 & 30.5 & 50.1  \\ \hline
NaVid~\cite{zhang2024navid}                  & \checkmark &    &    &   & 5.47 & 49.0 & 37.0 & 35.0 & - & -& -& -     \\
VLN-R1 ~\cite{vln_r1}                & \checkmark &    &    &   & 7.90 & 41.2 & 30.2 & 21.8 & 9.1   & 22.7 & 17.6 & -     \\
Uni-NaVid*~\cite{zhang2024uninavid}              & \checkmark &    &    &   & 5.58 & 53.3 & 47.0 & 42.7 & 6.24  & 48.7 & 40.9 & -     \\
NaVILA*~\cite{cheng2024navila}                 & \checkmark &    &    &   & 5.22 & 62.5 & 54.0 & 49.0 & 6.77  & 49.3 & 44.0 & 58.8  \\
StreamVLN~\cite{streamvln}              & \checkmark &    &    &   & 5.43 & 62.5 & 52.8 & 47.2 & 6.72  & 48.6 & 42.5 & 60.2  \\
DecoVLN (Ours)                   & \checkmark &    &    &   & \textbf{5.01} & \textbf{63.5} & \textbf{56.3} & \textbf{50.5} & \textbf{5.73}  & \textbf{54.2} & \textbf{46.3} & \textbf{63.5}  \\ \hline
\end{tabular}
\caption{Experimental results on the Val-Unseen dataset of R2R-CE and RxR-CE. * indicates training with additional large-scale datasets. Our method outperforms all approaches on the same benchmarks, even surpassing certain VLN models that were trained with additional large-scale datasets, without using global priors or multi-sensor inputs.}
\label{tab:main_results}
\vspace{-0.1cm}
\end{table*}

\cref{tab:main_results} summarizes our method’s performance on the R2R-CE and RxR-CE Val-Unseen benchmark datasets. The baseline models include both waypoint prediction models leveraging multi-sensor inputs and VLN models. 

Our method relies solely on ego-centric RGB inputs and surpasses all existing models on both benchmarks, achieving success rates of \textbf{56.3\%} and \textbf{54.2\%}, respectively. Compared to the state-of-the-art StreamVLN, our approach achieves \textbf{3.5\%} and \textbf{2.8\%} improvements in SR and SPL on R2R, and \textbf{5.6\%} and \textbf{3.8\%} improvements on RxR.
These significant gains strongly validate the superiority of our proposed \textbf{adaptive memory refinement} mechanism in handling long-horizon navigation tasks, effectively mitigating the information dilution problem commonly observed in traditional uniform sampling strategies.

It is also noteworthy that our model achieves these results \textit{without} any large-scale pretraining on datasets such as ScaleVLN~\cite{scalevln}, yet still outperforms most existing waypoint-based models that depend on multi-sensor fusion or global map priors. This demonstrates both the data efficiency and strong generalization capability of our framework. The results highlight that through carefully designed memory management mechanisms, the model can learn robust navigation policies from limited exposure, rather than relying purely on large-scale data accumulation.

\subsection{Ablation Studies}
To investigate the contribution of each module to the overall model performance, \cref{tab:module_abl} presents the results of the ablation experiments under different module configurations. The results demonstrate that each component yields a significant performance gain compared to the baseline model with uniform historical sampling. Introducing the adaptive memory refinement mechanism leads to a substantial improvement of \textbf{+3.6\%} in SR and \textbf{+2.2\%} in SPL, indicating that a higher contextual density effectively enhances the accuracy of action decision-making. Moreover, the \textbf{corrective fine-tuning} stage enables the model to perform semantic-aware error correction based on relevant historical observations, alleviating the issue of compounding errors in long-horizon navigation. As a result, SR improves by \textbf{+9.0\%}, while NE drops by \textbf{0.88\%} compared to the baseline.

\begin{table}[]
\centering
\begin{tabular}{p{1cm}p{1cm}cccc}
%\begin{tabular}{cccccc}
\hline
AMR & CF & NE ↓ & OS↑  & SR↑  & SPL↑ \\ \hline
   &           & 5.89 & 51.7 & 47.3 & 43.9 \\
\checkmark  &      & 5.50  & 57.9 & 50.9 & 46.1 \\
\checkmark  & \checkmark      & 5.01 & 63.5 & 56.3 & 50.5 \\ \hline
\end{tabular}
\caption{Ablation study of different modules on R2R Val-Unseen. AMR denotes Adaptive Memory Refinement, and CF denotes Corrective Fine-tuning.}
\label{tab:module_abl}
\vspace{-0.35cm}
\end{table}

\begin{figure}[t]
  \centering
   \includegraphics[width=1.0\linewidth]{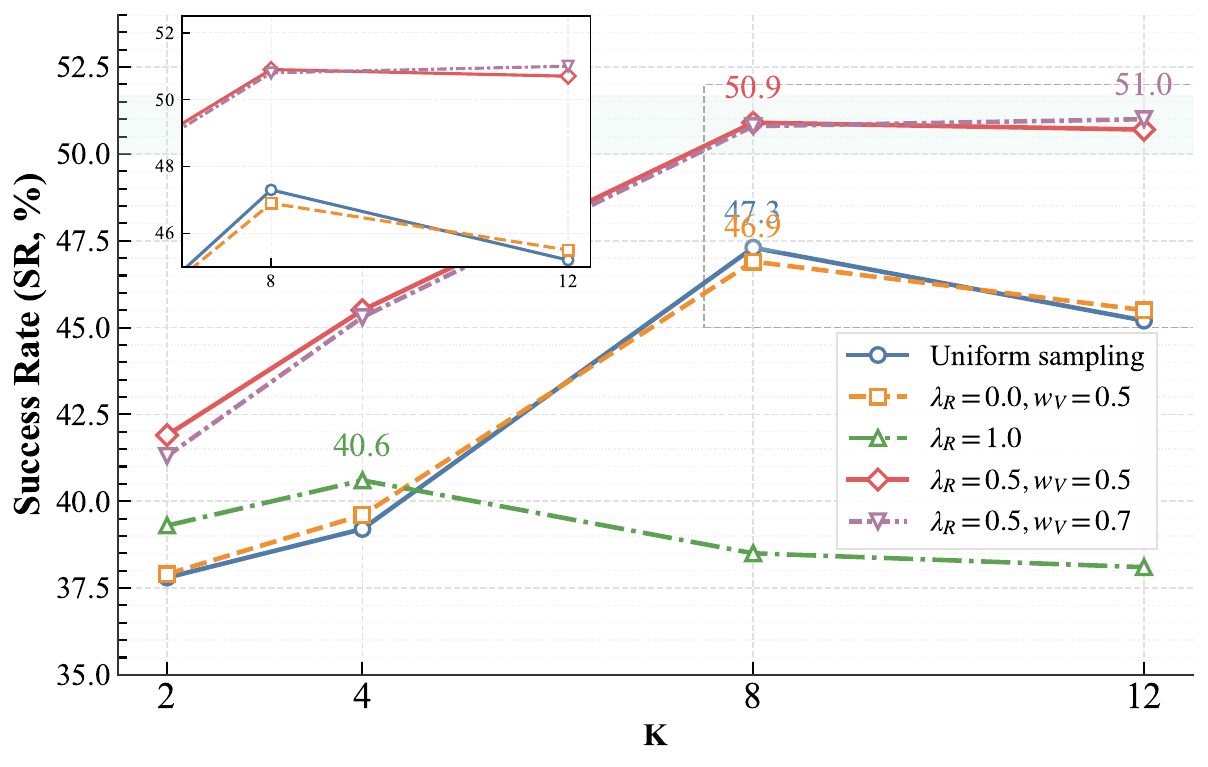}
   \caption{Evaluate success rates for various $K$ values and hyper-parameters on R2R Val-Unseen.}
   \label{fig:k_param}
   \vspace{-0.3cm}
\end{figure}

\textbf{Adaptive Memory Refinement (AMR).} To further analyze the effectiveness of the adaptive memory refinement mechanism,~\cref{fig:k_param} and~\cref{tab:k_param} illustrate the effects of different memory weights and memory sequence lengths $K$ on model performance without corrective fine-tuning. The results show that when $\lambda_R \rightarrow 1$, the selected memory sequence tends to focus on semantically similar consecutive frames with rich scene information. In this case, model performance is inversely correlated with sequence length and may even fall below that of uniform sampling, suggesting that redundant contextual semantics can harm long-term reasoning capability. Conversely, when $\lambda_R \rightarrow 0$, the model gradually degenerates into a uniform historical sampling strategy. 
The parameters $w_T$ and $w_V$ only have a minor impact on the final performance, though slightly increasing $w_V$ can improve navigation in complex environments requiring frequent turns, as it emphasizes visual diversity. 

\begin{table}[]
\centering
\begin{tabular}{ccccccc}
\hline
\multirow{2}{*}{K} & \multicolumn{3}{c}{R2R} & \multicolumn{3}{c}{RxR} \\ \cline{2-7} 
                   & NE ↓   & SR↑    & SPL↑  & NE ↓   & SR↑    & SPL↑  \\ \hline
2                  & 6.01   & 41.9   & 39.8  & 7.09   & 38.7   & 36.5  \\
4                  & 5.92   & 45.5   & 42.2  & 6.92   & 43.8   & 41.5  \\
8                  & \textbf{5.50}   & \textbf{50.9}   & \textbf{46.1}  & 6.53   & 49.2   & \textbf{45.2}  \\
12                 & 5.56   & 50.7   & 45.8  & \textbf{6.50}   & \textbf{49.5}   & 45.1  \\ \hline
\end{tabular}
\vspace{-0.1cm}
\caption{Ablation study of different historical memory lengths.}
\label{tab:k_param}
\vspace{-0.1cm}
\end{table}

Experimental results show that when the historical memory length does not exceed $K=8$, richer historical information can significantly improve navigation efficiency. However, as the memory capacity continues to increase, the uniformly sampled historical sequence tends to contain a large amount of redundant semantics, preventing the VLN model from effectively extracting key information for scene reasoning and causing a noticeable drop in performance. In contrast, after introducing the Adaptive Memory Refinement module, the model can more effectively utilize high-information-density state representations for decision-making, while alleviating the issue of contextual contamination caused by long memory sequences. 

\begin{figure}[]
  \centering
   \includegraphics[width=1.0\linewidth]{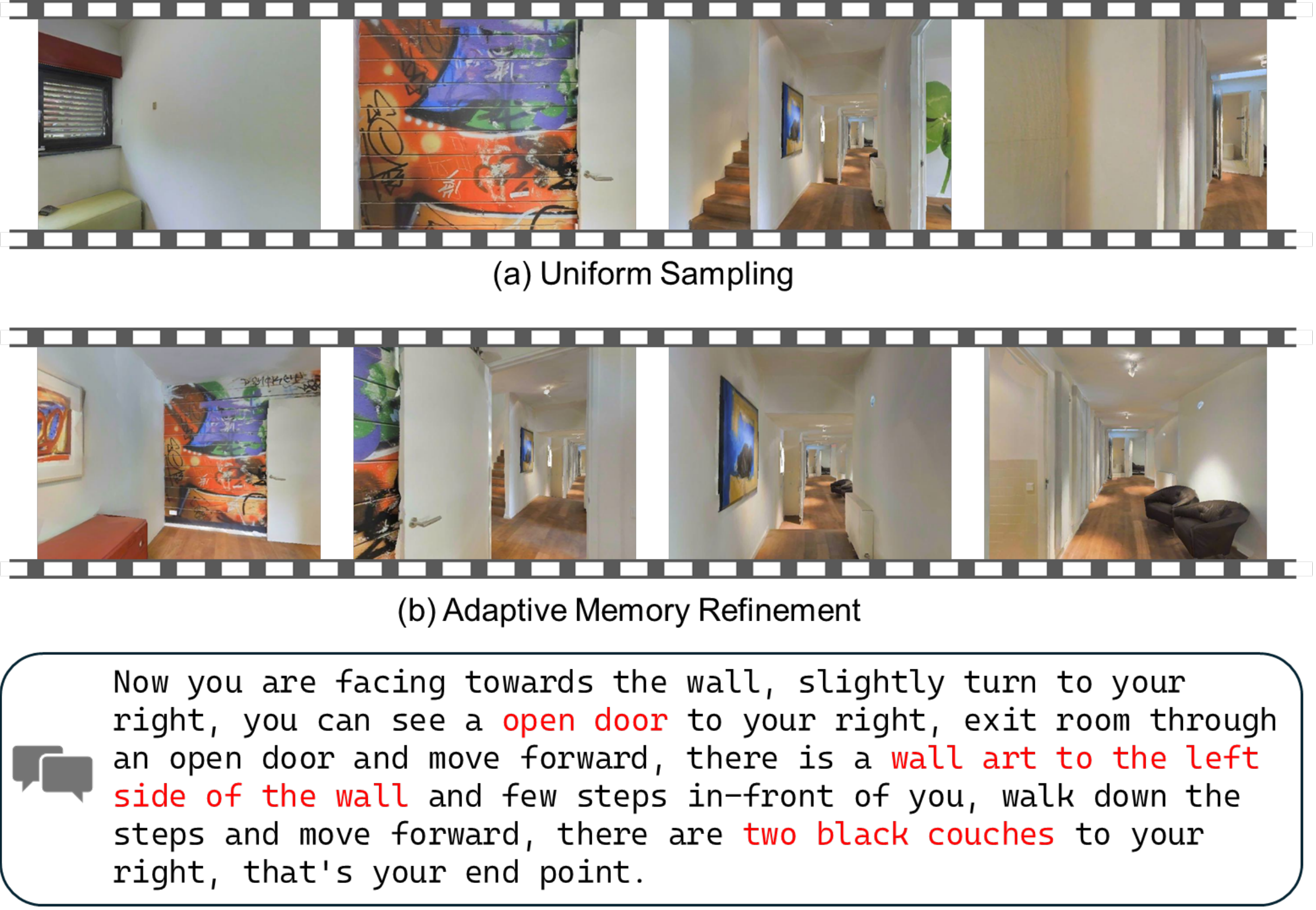}
   \vspace{-4mm}
   \caption{Comparison of Uniform Sampling strategy and Adaptive Memory Refinement mechanism. The history obtained via the uniform sampling strategy captures a large number of instruction-irrelevant images, such as walls and corners. This irrelevant semantic information severely impacts the model's reasoning performance. In contrast, the Adaptive Memory Refinement mechanism can effectively extract the key navigation points (indicated in \textcolor{red}{red}) from the instruction to achieve efficient and accurate navigation.}
   \label{fig:vis_amf}
   \vspace{-4mm}
\end{figure}

Although longer histories can in principle provide more comprehensive contextual cues for navigation planning, they also introduce greater inference latency and GPU memory consumption. Since navigation trajectories in the R2R and RxR benchmark datasets are typically between 10 and 15 meters in length, setting the memory size to $K=8$ achieves a favorable balance between navigation performance and computational efficiency.
We visualized the two sampling methods in \cref{fig:vis_amf}.

\textbf{Corrective Fine-tuning.} To further investigate the influence of the trusted region size on the effectiveness of corrective fine-tuning, we conducted an ablation study, as summarized in Table~\ref{tab:cft}. When the trusted region is set too small, the agent’s navigation policy becomes overly conservative. The model exhibits excessive sensitivity to minor deviations, often terminating navigation prematurely before substantial drift from the expert trajectory occurs, thereby limiting its exploration capacity. Conversely, if the trusted region is overly expanded, the agent may accumulate deviations that exceed its intrinsic corrective capacity during exploration. This not only introduces a considerable amount of low-quality corrective data and increases training cost but may also negatively impact the final learning efficiency. Empirical results indicate that a moderate trust-region threshold of $\tau = 3$ achieves the best balance between exploration robustness and corrective accuracy. Meanwhile, we compare our algorithm with the traditional DAgger algorithm in Table~\ref{tab:compare_dagger}, demonstrating its effectiveness.

\begin{table}[]
\centering
\begin{tabular}{ccccccc}
\hline
\multirow{2}{*}{$\tau$} & \multicolumn{3}{c}{R2R} & \multicolumn{3}{c}{RxR} \\ \cline{2-7} 
                        & NE ↓   & SR↑    & SPL↑  & NE ↓   & SR↑    & SPL↑  \\ \hline
1                       & 5.32   & 53.7   & 48.1  & 5.98   & 52.3   & 45.5  \\
3                       & \textbf{5.01}   & \textbf{56.1}   & \textbf{50.5}  & \textbf{5.73}   & \textbf{54.2}   & \textbf{46.3}  \\
6                       & 5.12   & 55.9   & 50.2  & 5.71   & 54.3   & 46.2  \\ \hline
\end{tabular}
\vspace{-0.1cm}
\caption{Ablation study of different trusted region sizes.}
\label{tab:cft}
\vspace{-0.3cm}
\end{table}

\begin{table}[t]
\centering
\begin{tabular}{lccccc}
\hline
Method  & Data & NE ↓ & OS ↑ & SR ↑ & SPL ↑ \\ \hline
Baseline & -    & 5.50 & 57.9 & 50.9 & 46.1  \\
DAgger  & 240K & 5.11 & \textbf{63.8} & 56.0 & 50.3  \\ 
DAgger  & 180K & 5.32 & 62.1 & 54.8 & 49.6  \\
% \rowcolor[HTML]{EAFFFE} 
Ours    & \textbf{180K} & \textbf{5.01} & 63.5 & \textbf{56.3} & \textbf{50.5}  \\ \hline
\end{tabular}
\vspace{-0.1cm}
\caption{Comparison with vanilla DAgger at different data scales.}
\label{tab:compare_dagger}
\vspace{-0.2cm}
\end{table}

\begin{figure*}[]
  \centering
   \includegraphics[width=1.0\linewidth]{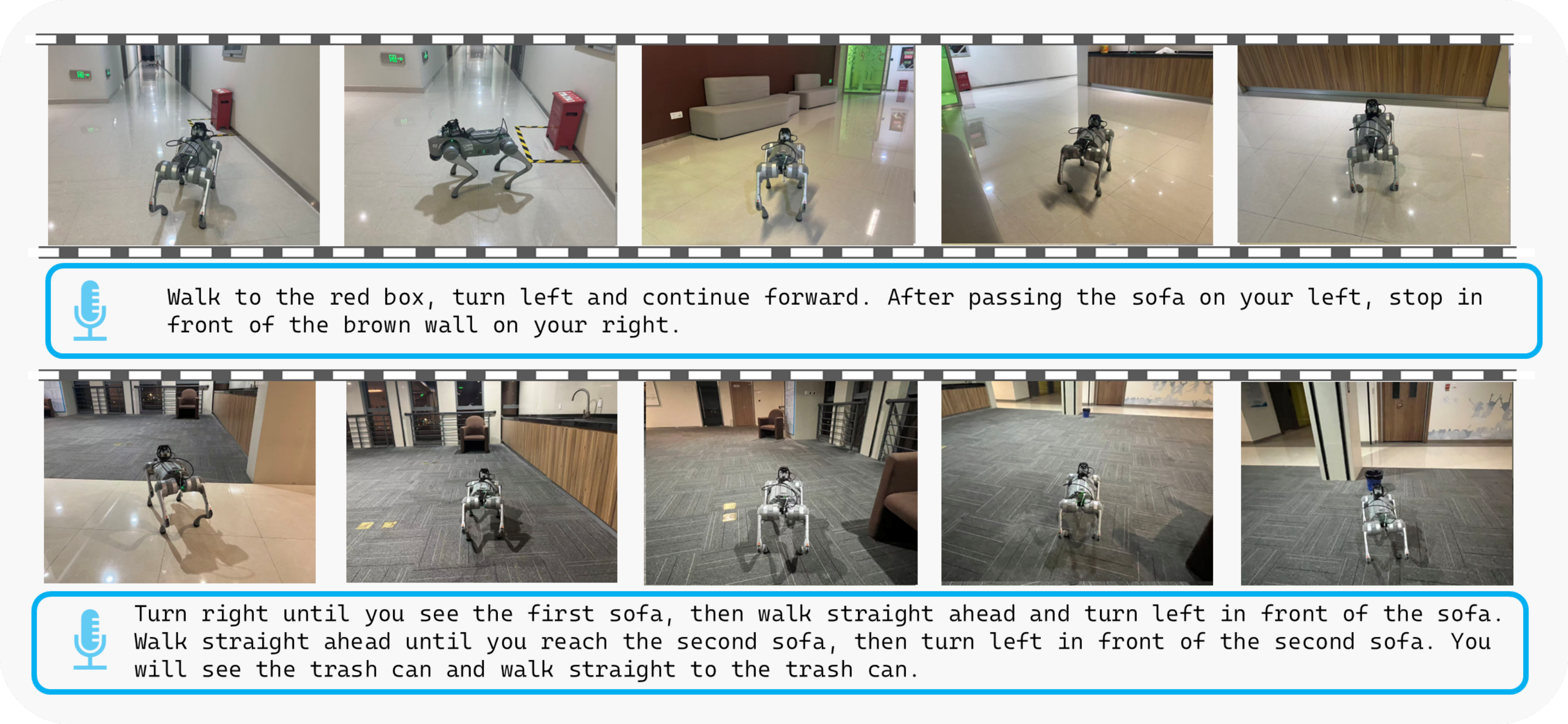}
   \caption{
   The robot accurately follows complex natural language instructions involving spatial reasoning and object grounding, demonstrating robust performance and strong sim-to-real generalization under challenging conditions.
   }
   \label{fig:real}
   \vspace{-1mm}
\end{figure*}

\subsection{Long Planning Horizon}

To rigorously evaluate the model's robustness in long-horizon scenarios, we construct a dedicated Long-horizon Navigation Validation Set based on the R2R and RxR benchmarks. The construction procedure is as follows: First, we filter trajectories exceeding 18 m in length from the original validation sets. Second, within the same environment, we select two independent trajectories whose endpoint and starting point are spatially proximate, and concatenated them using Trajectory Stitching to synthesize longer navigation paths. Finally, we utilize Qwen3~\cite{qwen3} to semantically fuse and rewrite the two original text instructions, generating a single, coherent long-text instruction. This validation set comprises a total of 536 long-horizon trajectories, with an average path length of 23 m.

\begin{figure}[]
  \centering
   \includegraphics[width=1.0\linewidth]{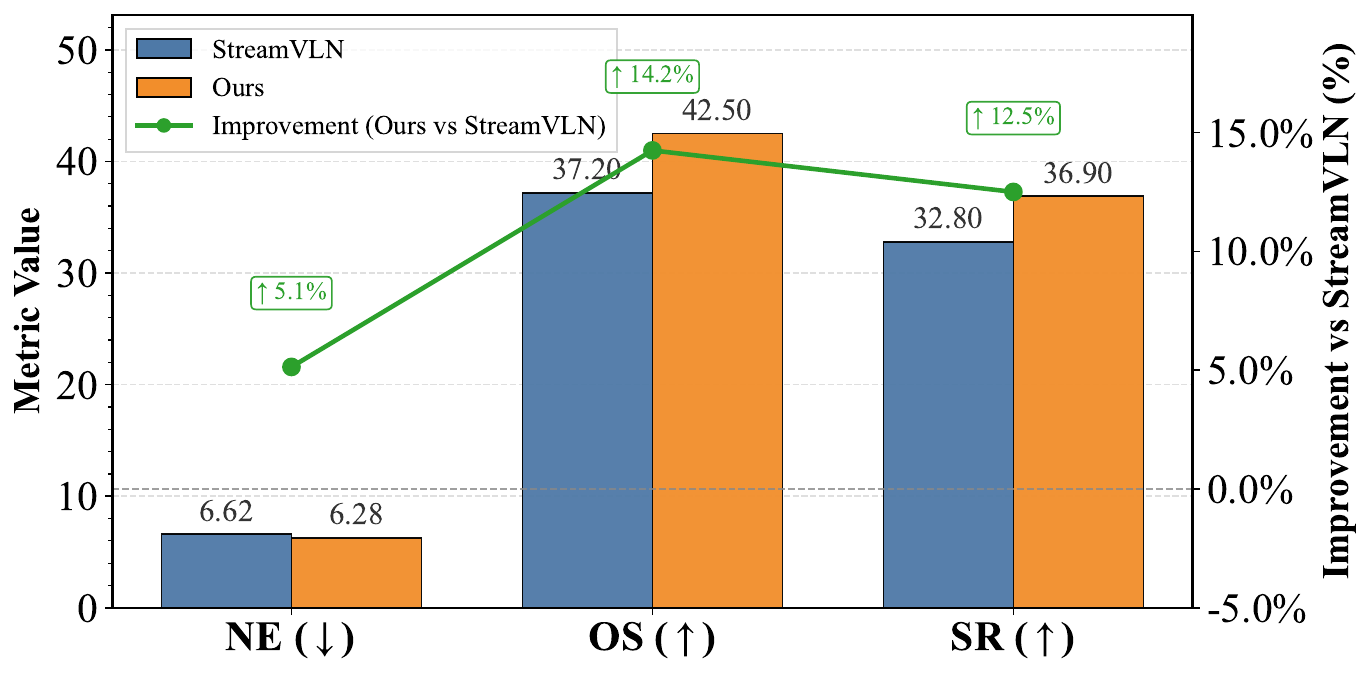}
   \caption{Performance comparison on long-horizon navigation validation set.}
   \label{fig:tra}
   \vspace{-5mm}
\end{figure}

The experimental results are presented in \cref{fig:tra}. The long planning horizon poses severe challenges to the VLN model's spatial memory retention and online error-correction capabilities. Quantitative analysis indicates that our adaptive memory refinement mechanism and Error-Correction Fine-tuning strategy effectively mitigate the problems of Catastrophic Forgetting and Compounding Errors, which are common in long-sequence decision-making. Notably, despite being trained only on standard short-horizon datasets, the model demonstrated exceptional navigation performance on the long-horizon validation set. This strongly proves the model's excellent generalization capability for long-horizon navigation tasks.

\subsection{Real-World Experiments}

We employ the Unitree GO2 as our quadruped robot platform, which transmits the real-time ego-centric video stream to a server equipped with a single RTX4090 GPU via a video transmission protocol. The robot utilizes a Jetson Orin as its core control unit, where a deployed Automatic Speech Recognition (ASR) model~\cite{jetson_voice} translates voice commands into text instructions and uploads them to the server. Subsequently, the server generates action chunks, each containing four distinct actions, and transmits them back to the control unit. The onboard controller parses these chunks into smooth motion trajectories and converts them into precise motor control signals via low-level motion APIs. All model training was conducted exclusively within the Habitat simulator, with no fine-tuning performed using real-world data. As illustrated in the ~\cref{fig:real}, we evaluated the model's performance in a real-world office environment characterized by complex layouts. Experimental results demonstrate that despite the significant domain gap between the simulator and the real world—regarding textures, lighting, and physical dynamics—our agent successfully comprehends instructions and plans reasonable paths within unknown physical spaces. Even when facing challenges difficult to model in simulations, such as floor reflections prevalent in real-world scenarios, the model consistently maintained the correct navigation direction. Additionally, 
we observed a significant emergent behavior: while moving toward the target point, the quadruped robot performs continuous lateral micro-adjustments and pose corrections to actively ensure that critical navigation waypoints or the final goal remain constantly within its field of view. This behavior demonstrates that the model is not merely executing pre-determined actions in an open-loop manner; rather, it possesses a degree of closed-loop feedback and anti-interference capabilities, thereby exhibiting strong robustness amidst complex real-world dynamics.

\section{Conclusion}
\label{sec:conclusion}

We present DecoVLN, a VLN framework whose core innovation lies in decoupling the agent’s observation, reasoning, and correction processes.
This design allows the agent to gather observations continuously while simultaneously executing actions and reasoning, effectively circumventing the high-latency bottlenecks inherent in the auto-regressive generation of VLMs. Without relying on any global prior information or depth sensors, DecoVLN achieves robust navigation in unknown environments using only ego-centric RGB inputs. To address the challenges prevalent in long-sequence navigation tasks, such as sparse semantic information, context pollution, and compounding errors, 
we propose an adaptive memory refinement mechanism and a state-action pair-based error-correction fine-tuning strategy. 
Our approach achieves superior performance on both the R2R and RxR benchmarks. Crucially, these results are achieved without relying on the additional large-scale ScaleVLN dataset, fully demonstrating DecoVLN's exceptional data efficiency and generalization potential.
{
    \small
    \bibliographystyle{ieeenat_fullname}
    \bibliography{main}
}

\clearpage

% \clearpage
\setcounter{section}{0}
\renewcommand{\thesection}
{\Alph{section}}
% \setcounter{page}{1}

% \maketitlesupplementary

\section*{Appendix}

\section{More Analysis on Observation-Reasoning Decoupling}
\subsection{Limitations of Existing Streaming Paradigms}
While existing Streaming VLN paradigms ~\cite{vln_r1, streamvln} achieve continuous environmental observation, their history management strategies exhibit critical efficiency limitations. These methods typically store all received observation frames into an exhaustive memory buffer, from which a fixed number of frames are uniformly sampled to construct the VLM’s input context during autoregressive action prediction. 
% When autoregressively generating action instructions, they then perform uniform sampling from this buffer to construct the VLM's context. 
This \textit{store-first, sample-later} pattern leads to three key problems: (1) \textbf{Context Pollution}: Uniform sampling is a task-agnostic strategy that completely ignores the semantic relevance between observation frames and the navigation task. Consequently, a large number of task-irrelevant, redundant frames are erroneously introduced into the context, severely interfering with the model's high-level reasoning and decision-making. (2) \textbf{Storage Overhead}: Indiscriminately caching all historical frames in memory causes storage occupancy to grow linearly with the number of navigation steps,  
% which is unsustainable for long-horizon navigation tasks.
rendering it inefficient and unsustainable for long-horizon navigation. (3) \textbf{I/O Bottleneck}: 
Before each autoregressive inference step, all frames must be re-sampled and transferred from memory (RAM) to GPU memory (VRAM), incurring additional data transfer latency. 
% At each reasoning step, the model must re-sample and transfer frames from RAM to GPU memory (VRAM), introducing non-negligible latency.
This undermines the supposed decoupling between perception and reasoning, as the model is forced to reprocess the entire observation stream during each inference step.

\begin{figure}[]
  \centering
   \includegraphics[width=1.0\linewidth]{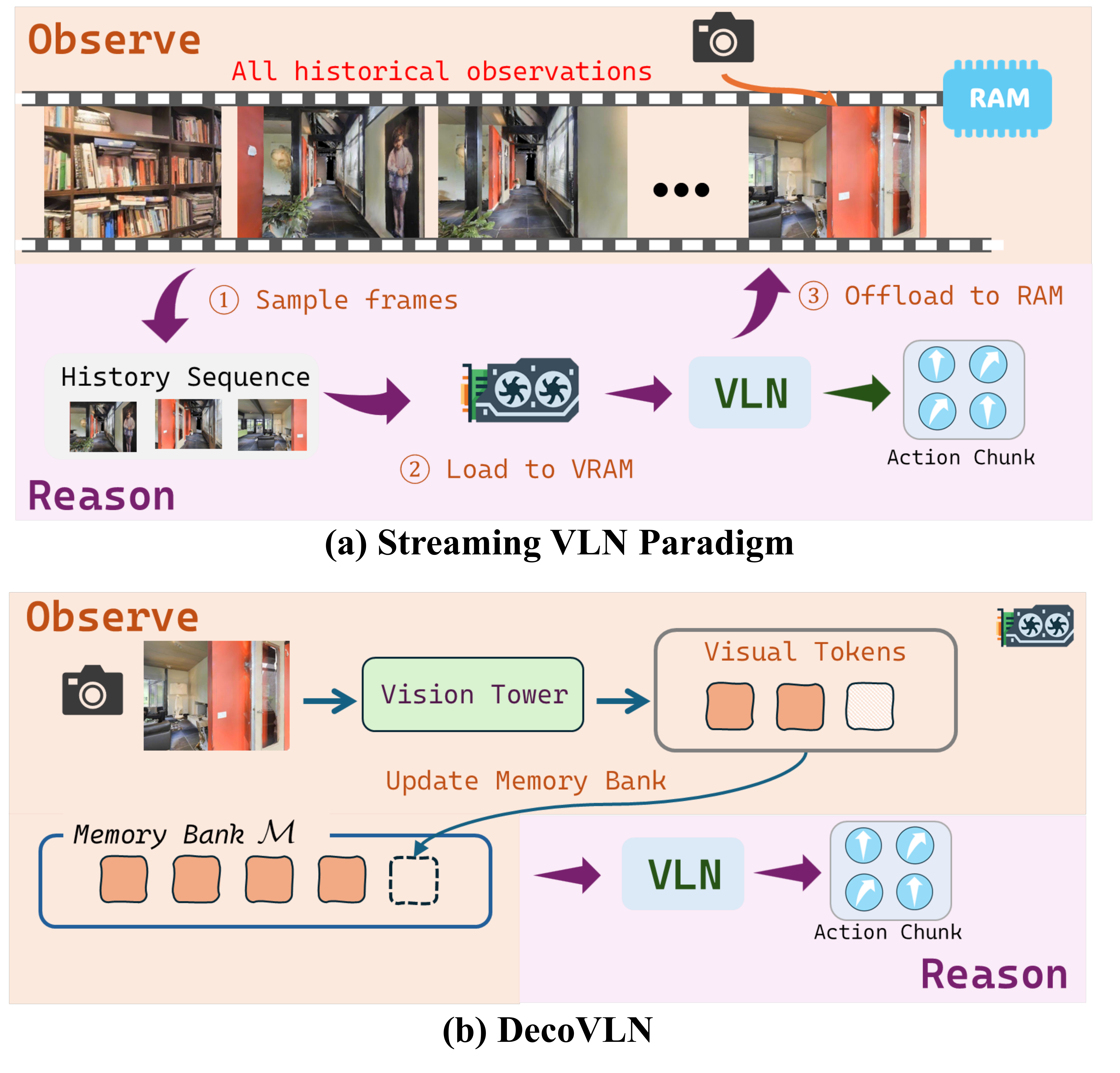}
   \caption{Comparison between the pipelines of Streaming VLN and DecoVLN. (a) Streaming VLN paradigm requires storing all historical observation sequences,
   sampling them during inference, and repeatedly transferring the selected frames between RAM and VRAM.
   % which must then be sampled from and transferred to VRAM for computation during the inference phase.
   This tight coupling of observation and reasoning leads to high storage overhead and inefficient data transfer.
   (b) Our DecoVLN, in contrast, introduces an adaptive memory-refinement mechanism during the observation phase.
   This design selectively preserves high-value semantic information in a VRAM-resident memory bank, which is directly consumed by the VLN model during inference.
   % to filter and for high-value semantic information. It dynamically updates a VRAM-resident memory bank, which is then directly input into the model during the inference phase.
   Our approach significantly reduces storage overhead and improves data transfer efficiency.}
   \label{fig:compare_stream}
   \vspace{-2mm}
\end{figure}

\subsection{Comparative Advantages of DecoVLN}
% Therefore, these methods do not truly achieve observation-reasoning decoupling. 
% the VLN model must re-filter the historical frame sequence for inference before every reasoning step.
% \cref{fig:compare_stream} compares the differences between the Streaming VLN paradigm and the DecoVLN. 
\cref{fig:compare_stream} highlights the key differences between conventional Streaming VLN paradigms~\cite{streamvln, zhang2024uninavid} and our proposed DecoVLN framework. DecoVLN achieves efficient decoupling of observation and reasoning. Firstly, the model directly calculates the relevance between frames and the instruction, storing only those frames strongly correlated with the navigation task as history. 
% Concurrently, by adding a frame similarity penalty term and a temporal factor, it prevents the model from focusing excessively on specific consecutive frames. 
To further promote diversity and temporal coverage, a visual similarity penalty and a temporal dispersion factor are introduced, preventing the memory from being dominated by redundant or consecutive frames.  The model's history queue is dynamically updated, and the entire queue remains resident in GPU memory (VRAM), significantly reducing both storage overhead and I/O latency.
% The latency in VLN models primarily stems from the token-by-token autoregressive generation method. After efficient decoupling, the framework allows the autoregressive generation phase to directly input the history queue containing visual tokens for inference, thereby greatly improving generation efficiency.
Since most latency in VLN models stems from token-by-token autoregressive generation, DecoVLN enables the model to directly access a compact, high-utility memory queue during inference, substantially improving reasoning efficiency without reprocessing the entire observation history.

\section{Additional Implementation Details}

\subsection{Dataset and Training Details}
To train a navigation model capable of handling the complexities of the physical world, we construct a simulation-based training dataset incorporating diverse challenges. The dataset generation is organized into two stages:  Supervised Fine-Tuning (SFT)  and Error-Correction Fine-Tuning (ECF).

As in mainstream VLMs~\cite{llava,qwen2_vl, li2025tokenpacker, yuan2024osprey,shi2024eagle,li2025eagle,yuan2025videorefer,yuan2025pixelrefer,yu2025inst3d,wang2025videoitg},  we leverage the SFT stage to establish strong cross-modal alignment between navigation instructions and expert trajectories.  We collect training data in the Habitat simulator from three datasets: R2R-CE~\cite{r2r}, R2R-EnvDrop~\cite{tan2019learning}, and RxR-CE~\cite{rxr}. To bridge the gap between ideal simulator motion and the inherent uncertainty of real-world actuators, we employ a stochastic step strategy during data collection. In real-world scenarios, factors such as wheel slip and limited motor precision introduce small deviations at every movement step, causing gradual trajectory drift. To mimic this phenomenon, we introduce slight random offsets into motion actions during data collection, improving policy's robustness to real-world noise and drift.
% It cannot achieve the perfect motion control seen in the simulator, and these errors accumulate, leading to trajectory divergence. Therefore, when collecting the simulation dataset, we introduce slight random offsets to the actions, mimicking the stochasticity of the physical world, thereby enhancing the policy's robustness.
To further enhance visual diversity, we apply several image augmentation strategies such as Gaussian blur and random masking, with  parameters listed in ~\cref{tab:augmentation_params}. 
Additionally, we employ a reverse augmentation strategy to expand the navigation paths. Specifically, we utilize the Qwen3 LLM~\cite{qwen3} to rewrite each original instruction $I$ into a reverse navigation instruction $I_{rev}$ representing the same path in the opposite direction.

During the ECF stage, the SFT-trained model $\pi_{SFT}$ performs autonomous rollouts in the simulator environment.  When the model deviates from the  expert trajectory, we introduce a \textit{ShortestPathFollower} (SPF) as an expert policy to provide  corrective actions that guide the agent back toward the optimal path. By collecting these high-quality state-action pairs, the model learns  to recover effectively from errors. Notably, during this phase, we  retain only visual-domain randomization and remove the action-level stochastic offsets to ensure the accuracy and stability of the expert correction signals.

\begin{table}[t]
\centering
\small
\begin{tabular*}{0.95\linewidth}{@{\extracolsep{\fill}} l c}
\toprule
\textbf{Augmentation Strategy} & \textbf{Parameters} \\
\midrule
Random Brightness    & factor = 0.2,\; prob = 0.5 \\
Random Saturation    & factor = 0.2,\; prob = 0.5 \\
Posterization        & bits = 4,\; prob = 0.5 \\
Random Sharpness     & factor = 0.2,\; prob = 0.5 \\
AutoContrast         & prob = 0.3 \\
Gaussian Blur        & prob = 0.2 \\
Random Masking       & prob = 0.2 \\
\bottomrule
\end{tabular*}
\caption{Image augmentation strategies and their corresponding parameters used in the SFT stage.}
\vspace{-1mm}
\label{tab:augmentation_params}
\end{table}

% % Table
% \begin{table}[]
% \centering
% \begin{tabular}{>{\centering\arraybackslash}p{4cm}>{\centering\arraybackslash}p{3cm}}
% \hline
% Augmentation Strategy          & Parameters         \\ \hline
% Random Brightness & f=0.2, p=0.5  \\
% Random Saturation & f=0.2, p=0.5  \\
% Posterization     & bits=4, p=0.5 \\
% Random Sharpness  & f=0.2, p=0.5  \\
% Autocontrast      & p=0.3         \\
% Gaussian Blur     & p=0.2         \\
% Random Masking       & p=0.2         \\ \hline
% \end{tabular}
% \caption{Parameters of data augmentation  for supervised fine-tuning.}
% \label{tab:details}
% % \vspace{-0.5cm}
% \end{table}

\subsection{Navigation Instructions}
To help the VLM to effectively understand its role within the POMDP~\cite{pomdp} and differentiate between information modalities, we adopted a structured prompt format, explicitly categorizing the input information into three types: System Message, Task Instruction, and Visual Context. First, the System Message is used to define the agent's role, with the relevant prompt being,``You are a robot designed for navigation tasks," which activates reasoning capabilities related to embodied tasks. Next, the Task Instruction includes the dataset’s natural language instruction $I$ along with pre-defined action space,  following standard VLN conventions: \textit{\{Move Forward 25cm, Turn Left 15 degrees, Turn Right 15 degrees, Stop\}}. Finally, the Visual Context is divided into two parts: historical memory and current observation. The historical memory consists of the refined sequence of historical frames obtained by the adaptive memory refinement strategy, while the current observation is the current ego-centric view. This explicit separation is designed to guide the model to prioritize anchoring its decisions on the current visual state, while treating the ``historical memory" as a queryable memory bank to assist with long-horizon reasoning and resolve perceptual aliasing issues.

\begin{table*}[t]
\centering
\small
\begin{tabular*} {0.93\linewidth}{@{\extracolsep{\fill}} l c |cccc |cccc}
\toprule
\multirow{2}{*}{\textbf{Method}} & \multirow{2}{*}{\textbf{Inputs}} 
& \multicolumn{4}{c|}{\textbf{R2R}} 
& \multicolumn{4}{c}{\textbf{RxR}} \\
\cmidrule(lr){3-6} \cmidrule(lr){7-10}
& & NE↓ & OS↑ & SR↑ & SPL↑ 
  & NE↓ & SR↑ & SPL↑ & nDTW↑ \\
\midrule
StreamVLN~\cite{streamvln} & RGB + Depth         & 4.98 & 64.2 & 56.9 & 51.9 & 6.22 & 52.9 & 46.0 & 61.9 \\
StreamVLN~\cite{streamvln} & RGB & 5.10 & 64.0 & 55.7 & 50.9 & 6.16 & 51.8 & 45.0 & 62.1 \\
\textbf{Ours}              & RGB         & 5.01 & 63.5 & 56.1 & 50.5 & 5.73 & 54.2 & 46.3 & 63.5 \\
\bottomrule
\end{tabular*}
\caption{Comparison with StreamVLN~\cite{streamvln} on the R2R and RxR benchmarks. 
Our method achieves competitive performance on R2R and significantly outperforms both RGB and RGB+Depth variants of StreamVLN on RxR, despite using only RGB inputs.}
\label{tab:compare_streamvln}
\end{table*}

\begin{figure*}[]
  \centering
   \includegraphics[width=1.0\linewidth]{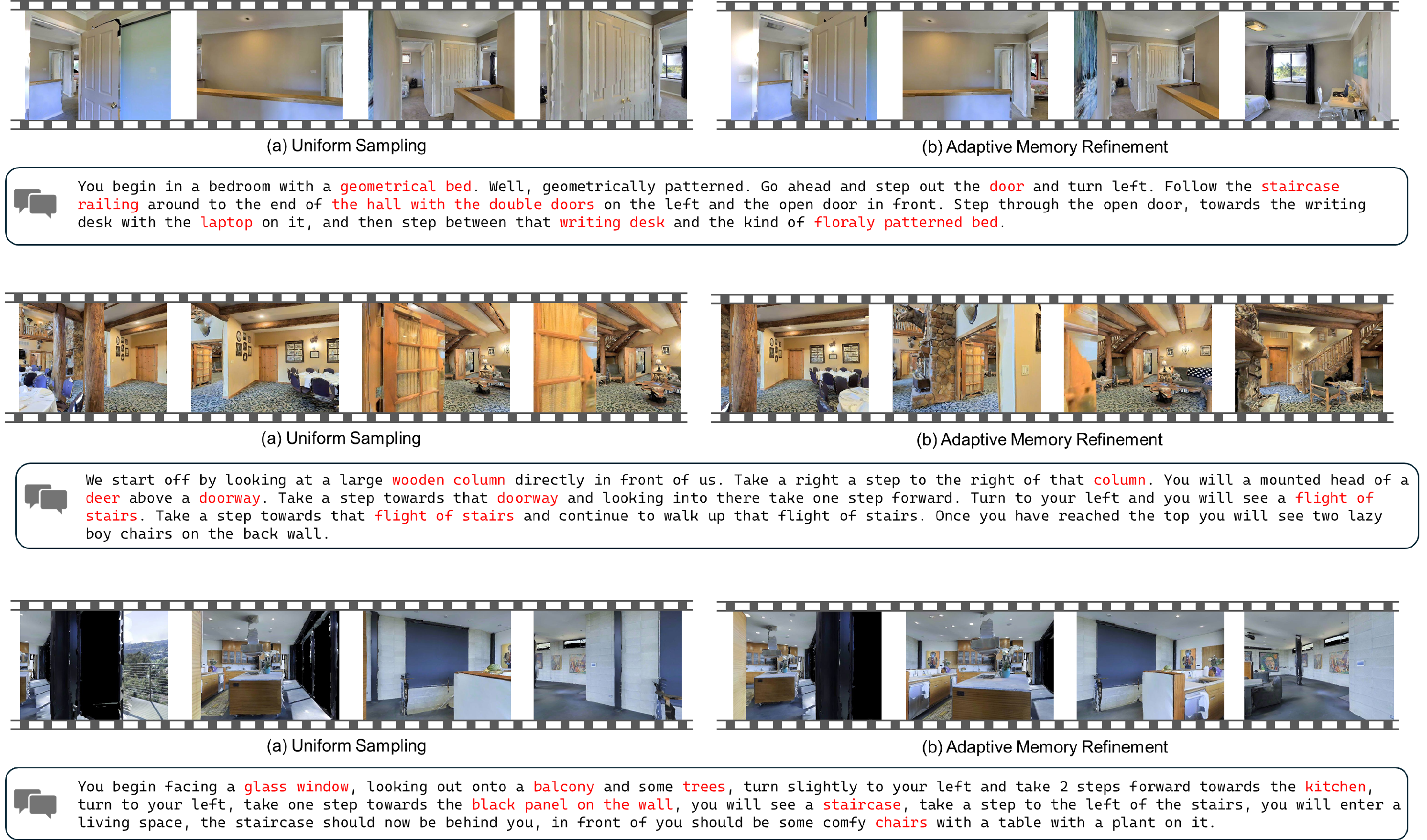}
   \caption{Visual comparison between uniform sampling and adaptive memory refinement. Frames selected by the proposed method exhibit higher information density and stronger semantic alignment with the navigation instructions, while effectively filtering redundant observations.}
   \label{fig:vis_am}
\end{figure*}

\section{More Discussions and Comparisons with Voxel-based Pruning}

\subsection{Detailed Discussions}
StreamVLN~\cite{streamvln} proposes a voxel-based spatial pruning strategy called the \textit{slow-updating memory context}. This method relies on depth maps computed from stereo cameras to project a uniformly sampled historical frame sequence into 3D space. It achieves semantic information compression by pruning redundant visual features in this 3D space, thereby improving navigation efficiency. However, this process not only introduces an additional dependency on depth sensors but also involves epipolar constraint calculations and 3D reconstruction, which significantly increases inference latency and limits navigation efficiency. In contrast, our adaptive memory refinement strategy only requires ego-centric RGB images.

More fundamentally, StreamVLN's voxel pruning is a \textbf{post-hoc} process: it operates on a task-agnostic, uniformly sampled historical sequence. This ``sample-first, prune-later" design cannot evaluate the semantic relevance of an observation frame to the navigation task at the source of information acquisition. Consequently, even highly redundant or instruction-irrelevant observations, such as blank walls, can be uniformly sampled and retained in the context. This prevents the model from improving the contextual signal-to-noise ratio at its root. In contrast, our adaptive memory refinement strategy is a task-aware, \textbf{pre-hoc} filtering mechanism. It assesses the relevance of an observation before it enters the memory queue, fundamentally guaranteeing a high information density in the context.

\subsection{Performance Comparisons}
\begin{figure*}[]
  \centering
   \includegraphics[width=1.0\linewidth]{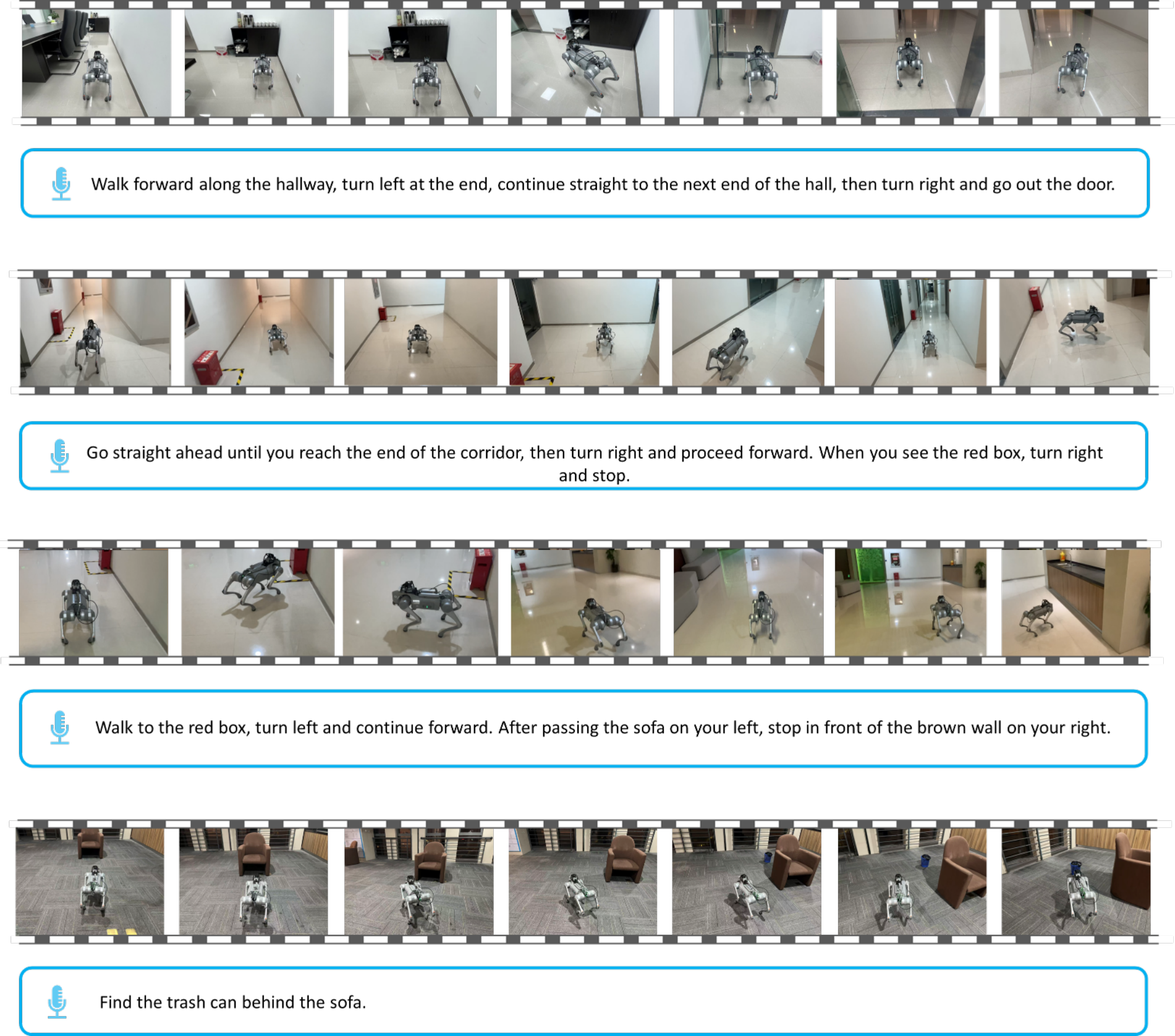}
   \caption{Real-world navigation results of DecoVLN across diverse  scenarios. The robot accurately follows complex natural language instructions involving spatial reasoning and object grounding, demonstrating robust performance and strong sim-to-real generalization under challenging conditions.}
   \label{fig:vis_real}
\end{figure*}

A comprehensive comparison with different versions of StreamVLN~\cite{streamvln} is shown in ~\cref{tab:compare_streamvln}. Our method comprehensively surpasses StreamVLN on the same benchmarks, and it even outperforms the depth-enabled version of StreamVLN on the RxR dataset. Furthermore, our model's dataset size and training time are significantly smaller than those of StreamVLN. This is because although the voxel-based pruning strategy increases semantic information density, the underlying uniform sampling strategy ignores the relevance of the images to the navigation task and may capture irrelevant images such as walls, failing to improve the contextual signal-to-noise ratio from the source. Extensive experiments are conducted across a wide range of diverse scenarios, as illustrated in \cref{fig:vis_real}.
The adaptive memory refinement module can reconstruct an effective historical trajectory using only sparse keyframes, enabling efficient long-horizon compression while significantly reducing inference overhead, thereby validating the decoupled observation–reasoning design.
Crucially, our model achieves the aforementioned results without utilizing the large-scale ScaleVLN~\cite{scalevln} dataset, while requiring significantly less training time compared to StreamVLN. This strongly corroborates the efficiency of DecoVLN in extracting effective features.

\section{Real-World Deployment}
\subsection{Platform Setup}
DecoVLN outputs symbolic text commands rather than precise, low-level control signals (e.g., joint torques).  
This design theoretically achieves \textbf{platform-agnosticism}, wherein the high-level VLM navigation logic is decoupled from the low-level, embodiment-specific motor control. 
Consequently, DecoVLN's high-level policy has the potential to be migrated to diverse hardware platforms, including quadruped, wheeled, and even humanoid robots. 

In our real-world experiments, we  deploy DecoVLN on a Unitree GO2 quadruped robot. The system adopts a device-server collaboration architecture. The robot is equipped with a Jetson Orin, which serves as the onboard computational core. It is responsible for running a lightweight Automatic Speech Recognition (ASR) model and the low-level motion control APIs. A remote server with a single NVIDIA RTX 4090 GPU runs the computationally intensive DecoVLN model. The system's asynchronous data loop operates as follows: the onboard ASR model converts the user's voice commands into text instructions $I$ in real-time. These instructions and the current ego-centric video stream $o_t$, are transmitted (\textit{uplinked}) to the server via a wireless video transmission protocol. Upon receiving this data, 
the server performs VLM inference and returns (\textit{downlinked}) the resulting symbolic action chunk, containing four discrete navigation actions, which is then executed by the onboard controller.

\subsection{Sim-to-Real Transfer}
\subsubsection{Bridging the Execution Gap}
To address the \textit{Execution Gap} between the simulation and real-world environments, and to enhance motion stability and efficiency, we avoid the naive strategy of executing discrete actions sequentially. On a dynamic quadruped platform, this approach leads to discontinuous motion and is highly prone to compounding errors. Instead, the onboard controller is designed to interpret the action chunk received from the server as a holistic short-horizon goal. Specifically, rather than executing these four actions serially, the controller directly computes the final relative pose ($\Delta x, \Delta y, \Delta \theta$) that corresponds to the entire action sequence. For example, an action chunk \textit{[forward 25cm, forward 25cm, turn left 15 degree, stop]} is parsed into a target state of ($\Delta x=0.5m, \Delta y=0, \Delta \theta=15°$). Subsequently, the onboard controller invokes the low-level motion APIs to compute and execute a smooth and dynamically feasible trajectory to approximate this target.  This conversion from discrete symbolic output to continuous motion substantially improves real-world navigation robustness and efficiency.
% This approach transforms the VLM's discrete text output into smooth, continuous control, significantly enhancing both navigation efficiency and robustness.

\subsubsection{Qualitative Real-world Results}
To validate the model's performance in bridging the sim-to-real gap, we deployed the model in a zero-shot manner within diverse complex real-world scenarios, with the test results shown in the ~\cref{fig:vis_real}. These qualitative examples highlight the model's capabilities in fine-grained instruction understanding and semantic reasoning. The model not only precisely executes basic motion instructions but also, when observing open-vocabulary objects that never appeared in the navigation training set, remains capable of correct identification and localization based on context. Furthermore, the model exhibits robustness against environmental disturbances. Real physical environments often encompass complex factors difficult to precisely model in simulators, such as dynamic lighting, specular reflections, and unstructured obstacles; yet, the model maintains stable perception and decision-making, planning paths that are both geometrically reasonable and semantically correct.

\section{Limitations and Future Work}
Our current VLN framework is  is built upon LLaVA-Video-7B,  
whose computational cost prevents fully onboard inference on devices such as the Jetson Orin. Current real-time operation relies on a device–server collaborative architecture, where heavy VLM inference is offloaded to a remote GPU server via wireless communication. This setup introduces additional network latency and limits deployment in environments with poor connectivity.
Future work will explore model distillation and parameter-efficient tuning to transfer the high-level navigation capabilities of the 7B model to more lightweight models ranging from 0.5B to 2B parameters,  enabling fully onboard real-time inference.

In challenging real-world scenes, the agent may still lose orientation when visual landmarks disappear or perceptual aliasing occurs.
The current system lacks a deep introspective mechanism to address such scenarios of total deviation.  To solve this problem, we plan to incorporate a Chain-of-Thought (CoT)–based error-recovery module that leverages historical memory to backtrack, identify a reliable waypoint, and replan a corrected trajectory, improving robustness in long-horizon navigation.

% WARNING: do not forget to delete the supplementary pages from your submission 
% \input{sec/X_suppl}

\end{document}